\documentclass[conference]{IEEEtran}
\IEEEoverridecommandlockouts
\usepackage{cite}
\usepackage{amsmath,amssymb,amsfonts}
\usepackage{algorithmic}
\usepackage{graphicx}
\usepackage{textcomp}
\usepackage{booktabs}
\PassOptionsToPackage{hyphens}{url}\usepackage[breaklinks=true,bookmarks=false]{hyperref}
\usepackage[table,xcdraw,rgb]{xcolor}
\usepackage{tikz}
\usepackage{nicefrac, xfrac}
\usepackage{mathtools}
\usepackage{array,multirow,graphicx}
\usepackage{pifont}
\def\BibTeX{{\rm B\kern-.05em{\sc i\kern-.025em b}\kern-.08em
    T\kern-.1667em\lower.7ex\hbox{E}\kern-.125emX}}
\usepackage{fancyhdr}

\newcommand{\yes}{\ding{51}}%
\newcommand{\skeleton}{\includegraphics[height=2.5mm]{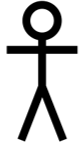}} 
\newcommand{\object}{\includegraphics[height=2.0mm]{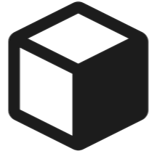}}
\newcommand{\skeletonAndObject}{\includegraphics[height=2.5mm]{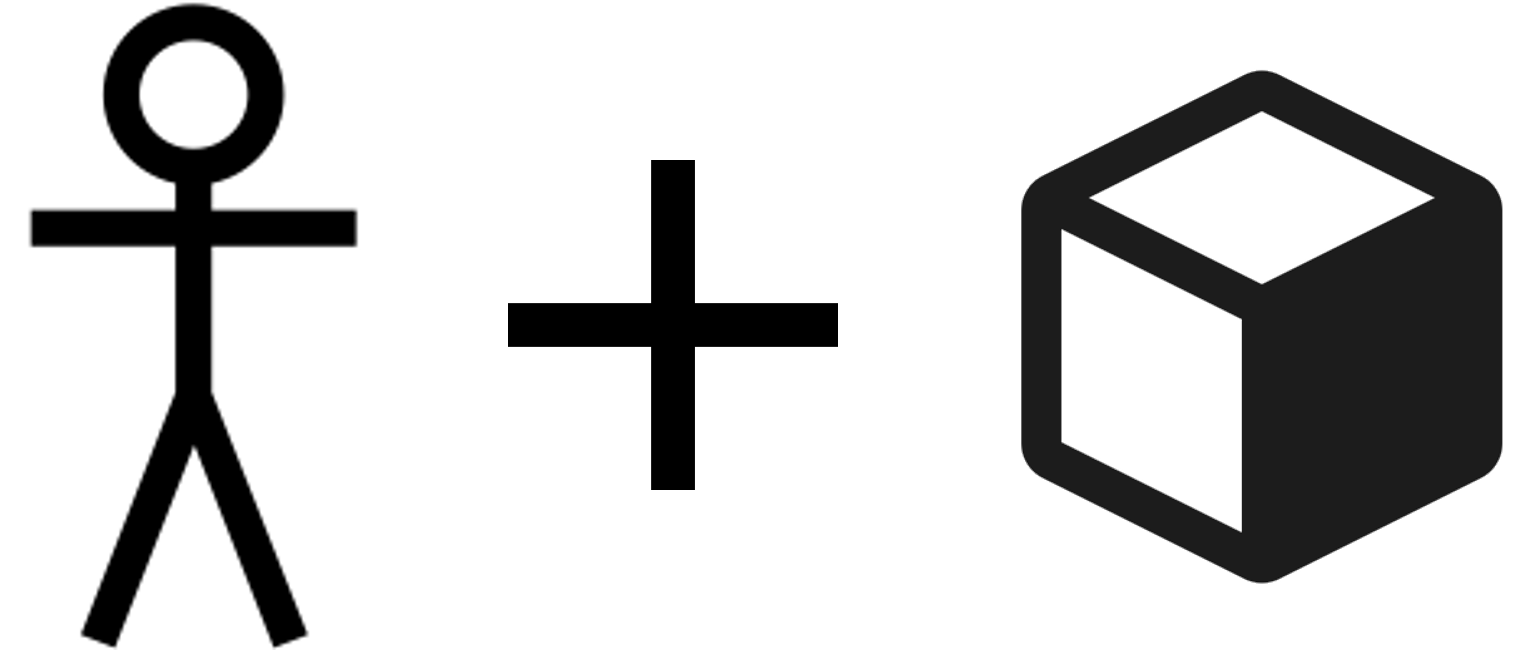}}
    
\begin{document}

\renewcommand*{\figureautorefname}{Fig.}
\renewcommand*{\tableautorefname}{Tab.}
\renewcommand*{\equationautorefname}{Eq.}
\renewcommand*{\sectionautorefname}{Sec.}
\renewcommand*{\subsectionautorefname}{Sec.}

\title{How Object Information Improves Skeleton-based\\ Human Action Recognition in Assembly Tasks\\
\thanks{This work has received funding from the Carl-Zeiss-Stiftung as part of the project engineering for smart manufacturing (E4SM).}
}

\author{\IEEEauthorblockN{Dustin Aganian, Mona Köhler, Sebastian Baake, Markus Eisenbach, and Horst-Michael Groß}
\IEEEauthorblockA{\textit{Ilmenau University of Technology, Neuroinformatics and Cognitive Robotics Lab} \\
98684 Ilmenau, Germany \\
\small\tt dustin.aganian@tu-ilmenau.de, ORCID: 0009-0006-3925-6718
}}

\maketitle

\newcommand\todo[1]{\textcolor{red}{{#1}}}
\newcommand\redtext[1]{\textcolor{red}{{#1}}}

\newboolean{isarxiv}
\setboolean{isarxiv}{true}
\ifthenelse{\boolean{isarxiv}}{%
    \renewcommand{\headrulewidth}{0pt}
    \fancypagestyle{fancyfirstpage}{%
        \fancyhf{}%
        \fancyhead[C]{%
            \scriptsize%
               \textcolor{lightgray}{%
                  \small%
                   This work has been submitted to the IEEE for possible publication.\\
                   Copyright may be transferred without notice, after which this version may no longer be accessible.%
               }%
        }
        \fancyfoot[C]{%
            \footnotesize%
            \textcolor{gray}{\thepage}%
        }
    }
    \fancypagestyle{fancypage}{%
        \fancyhf{}%
        \fancyfoot[C]{%
            \footnotesize%
            \textcolor{gray}{\thepage}%
        }
    }
    \thispagestyle{fancyfirstpage}
    \pagestyle{fancypage}
}{%
    \thispagestyle{empty}%
    \pagestyle{empty}%
}%

\begin{abstract}
As the use of collaborative robots (cobots) in industrial manufacturing continues to grow, human action recognition for effective human-robot collaboration becomes increasingly important.
This ability is crucial for cobots to act autonomously and assist in assembly tasks.
Recently, skeleton-based approaches are often used as they tend to generalize better to different people and environments.
However, when processing skeletons alone, information about the objects a human interacts with is lost.
Therefore, we present a novel approach of integrating object information into skeleton-based action recognition.
We enhance two state-of-the-art methods by treating object centers as further skeleton joints.
Our experiments on the assembly dataset IKEA ASM show that our approach improves the performance of these state-of-the-art methods to a large extent when combining skeleton joints with objects predicted by a state-of-the-art instance segmentation model. 
Our research sheds light on the benefits of combining skeleton joints with object information for human action recognition in assembly tasks.
We analyze the effect of the object detector on the combination for action classification and discuss the important factors that must be taken into account.

\end{abstract}

\section{Introduction}

In the course of industry 4.0, cooperation between humans and situation-aware collaborative robots (cobots) is becoming increasingly important for manufacturing processes~\cite{matheson2019human,sherwani2020collaborative,inkulu2021challenges}.
In order to effectively assist in assembly processes, cobots must first be able to visually perceive the worker and recognize the current assembly state~\cite{wang2019symbiotic,al2019action,li2021toward,eisenbach2021,fan2022vision,terreran2023skeleton}.
One crucial aspect of attaining this goal is human action recognition~\cite{eisenbach2021,terreran2023skeleton}.

\begin{figure}[!t]
    \centering
    \includegraphics[width=\linewidth]{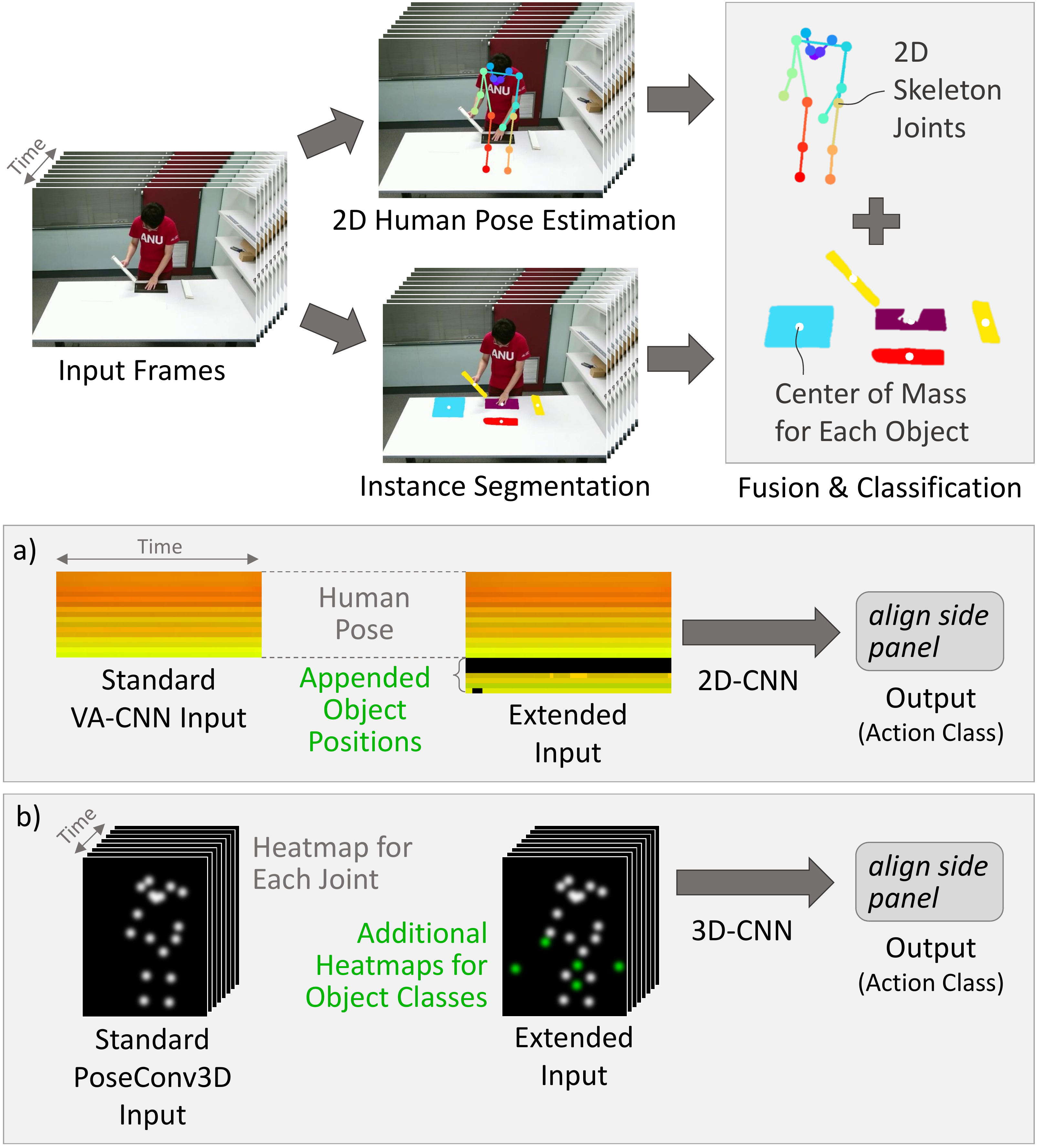}
    \caption{%
    Overview of our approach.
    We combine skeleton joints and object information for two human action recognition methods, namely a) VA-CNN~\cite{VACNN-TPAMI2019} and b) PoseConv3D~\cite{PoseConv3D-cvpr2022}.
    Note that, for PoseConv3D, we actually have one heatmap per joint and stack frames in an additional 4\textsuperscript{th} dimension.
    The different colors for object centers are used for visualization purposes only.
    Example frames taken from IKEA ASM dataset \cite{IKEA-wacv2021}.
    }%
    \vspace*{-2mm}
    \label{fig:eyecatcher}
\end{figure}

For human action recognition, often RGB-based approaches are used.
However, these approaches face several challenges in scenarios where the training dataset and the application environment do not have a significant overlap.
Due to the difficulties in creating action recognition datasets for assembly processes~\cite{IKEA-wacv2021}, these datasets tend to be smaller compared to other areas such as object detection~\cite{coco-eccv2014}.
As a result, RGB-based approaches trained on these datasets tend to overfit and exhibit poor generalization capabilities.
This can lead to difficulties when the person performing the action changes or when the background or working environment during deployment are drastically different to training.
To overcome these limitations, skeleton-based action recognition models are the better alternative, as they tend to generalize much better under such circumstances~\cite{terreran2023skeleton}.
Nevertheless, skeleton-based models have the drawback of not being able to process objects involved in actions, such as when a worker picks up a hammer or a screwdriver.
Similarly, actions in which workers pick up similar-sized assembly parts, such as a cabinet side panel or a cabinet back panel, cannot be distinguished based on skeletons.

In contrast to action recognition, for object detection there are highly diverse, public datasets~\cite{coco-eccv2014} available, which enable the training of very accurate and well generalizing object detectors~\cite{maskrcnn-iccv2017}.
Therefore, our goal in this paper is to develop a novel approach to improve human action recognition by incorporating additional object information into skeleton-based action recognition for trimmed sequences.

We will demonstrate that a relatively straightforward enhancement of the input encoding (as shown in \autoref{fig:eyecatcher}) can greatly improve the performance of two different methods.
Thereby we use the state-of-the-art dataset IKEA ASM~\cite{IKEA-wacv2021}, a comprehensive assembly dataset for action recognition with human-object interactions.
In order to analyze the performance gain of combining object information with skeleton information, we first experiment with ground truth object masks, which are independent of the accuracy of an object detector.
Afterwards, we examine the effect of using predicted objects masks and present the impact on the performance caused by noisier predictions.

Our contributions in this paper can be summarized as follows:
\begin{enumerate}
    \item We propose a novel approach to integrate object information into skeleton-based action recognition.
    \item Our experiments with two different methods for skeleton-based action recognition demonstrate the usefulness of combining skeletons and object positions.
    \item We show the influence of ground truth versus predicted objects on the performance of action recognition.
    \item Overall, our approach massively improves the performance of action recognition on the IKEA ASM dataset~\cite{IKEA-wacv2021}.
\end{enumerate}

\section{Related Work}

In this section, we will first present an overview of the current state of the art for skeleton-based human action recognition, and then specify the methods that we will modify for incorporating object information as well.
Afterwards, we will outline other work in the field of human action recognition that specifically incorporate object information.

\subsection{Skeleton-based Action Recognition}
The general term human action recognition encompasses a wide range of subfields.
In this paper, however, we want to focus on the action classification task of pre-trimmed video clips of human skeleton sequences.
This task serves as a foundation, since other methods for tasks like action segmentation or action detection, such as PDAN~\cite{Dai2021PDAN} or MS-TCT~\cite{Dai2022MS-TCT}, typically use models trained on action recognition of pre-trimmed clips as the backbone of their method.

Regarding skeleton-based action recognition, RNNs have been used for a long time~\cite{Presti2016SkeletonSurvey}.
Currently 2D CNNs such as VA-CNN~\cite{VACNN-TPAMI2019}, 3D CNNs like PoseConv3D~\cite{PoseConv3D-cvpr2022} and graph convolution networks (e.g. 2S-AGCN~\cite{Shi2019AGCN}, Si-GCN~\cite{liu2019si}) are typically used.

In this paper, we enhance the VA-CNN and PoseConv3D methods that both achieve state-of-the-art results on common skeleton-based action recognition benchmark datasets such as NTU~RGB+D~\cite{Shahroudy2016NTU}.
Furthermore, both methods allow for our modifications to incorporate object information sequences in addition to skeleton sequences.
Here, VA-CNN suited itself because it is able to train on very long or short clips easily, since it can process any length of clips due to its input coding.
Our preliminary studies have also shown that VA-CNN achieved better results than other methods we tested, such as 2S-AGCN.
Furthermore, we chose PoseConv3D because, unlike most other skeleton-based action recognition methods, it is able to easily handle multiple people in a frame and thus multiple instances of the same joint class.
Why this is advantageous and how the two selected methods VA-CNN and PoseConv3D as well as our modifications work in detail will be described in \autoref{section-approach}.

\subsection{Action Recognition Incorporating Object Information}
 Object information can play a crucial role in improving action recognition.
Research in this area typically focuses on advancing human action recognition.
Some other studies are done in the field of human-object interaction recognition, with the goal of recognizing actions, where the objects used are included in the action labels.

An example of such work is~\cite{xu2022skeleton} in which skeletons are linked with object position information.
However, this combination does not take the object class into account, nor can more than one object be processed.
In contrast, the presented method in~\cite{dreher2019learning} is not limited to the number of objects used.
Here, using a rather simple graph network on a very small dataset, the authors demonstrated that the combination of object information with skeleton information generally works to train an action classifier.
In the recently published work~\cite{xing2022trash}, object information is included in a graph convolutional network to perform action segmentation.
In their experiments, ground truth object boxes were used exclusively, which benefited the developed method.
However, in our experiments, we show, that the performance using ground truth objects does not resemble the real performance when using detected objects.
Furthermore, in~\cite{xing2022trash} neither the contribution of using the object boxes on the overall performance was shown, nor the use of detected objects was investigated.

In addition to these studies, there are works which investigate the use of object information through simulation.
In~\cite{zakour2021hoisim}, a method is introduced for simulating human actions and object movements to predict actions and activities.
For the prediction, RNNs and Transformers were employed, but the performance of the developed methods has not been assessed on real datasets.

In contrast to related work, our approach can handle multiple objects, while simultaneously encoding class information along with the object positions.
Furthermore, we show the general performance gain of combining object information with skeleton information compared to using solely skeleton information, an approach which has only been sparsely studied in the current state of the art.
Moreover, we demonstrate how our methods perform on a large dataset with real estimated object information instead of simply using ground truth object masks.

\section{Used dataset}
As previously described, our focus in this paper is to improve action recognition for cobots in the context of manufacturing processes.
Action recognition datasets that are closest to this scenario are typically assembly datasets.
Among them are datasets mostly recorded in first-person perspective or a very close surveillance perspective, such as Assembly101~\cite{assembly101-cvpr2022} or Meccano~\cite{meccano-wacv2021}.
Yet there are also datasets like IKEA ASM~\cite{IKEA-wacv2021} and ATTACH~\cite{attach}, which were recorded from a surveillance perspective, which is very similar to the perspective of a cobot that observes a worker in order to help out.

For our experiments, we have decided to use the furniture assembly dataset IKEA ASM, which is a very extensive dataset, captured from multiple perspectives.
It consists of scenes where different furniture types are assembled in various environments, making the dataset quite challenging.
Moreover, in contrast to ATTACH, it also contains labeled object masks.

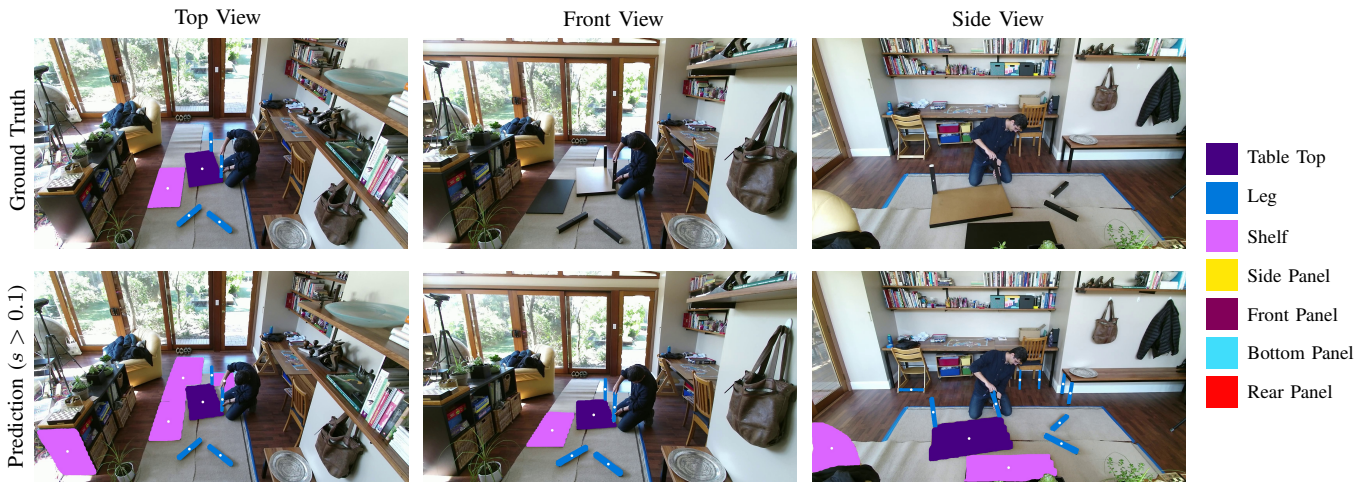
\begin{figure*}[htb]
    \centering
    \definecolor{Table Top}{RGB}{70,0,129}%
    \definecolor{Leg}{RGB}{0, 120,220}%
    \definecolor{Shelf}{RGB}{220,100,255}%
    \definecolor{Side Panel}{RGB}{255,231,6}%
    \definecolor{Front Panel}{RGB}{130,0,89}%
    \definecolor{Bottom Panel}{RGB}{64,221,251}%
    \definecolor{Rear Panel}{RGB}{255,5,5}%
    \resizebox{\linewidth}{!}{%
    \begin{tikzpicture}%
        \node[anchor=west, rotate=90] at (-0.1, -1){\footnotesize{Ground Truth}};%
        \node[anchor=west] at (1.8, 1.6){\footnotesize{Top View}};%
        \node[anchor=west] at (0, 0){%
	        \includegraphics[width=4.8cm]{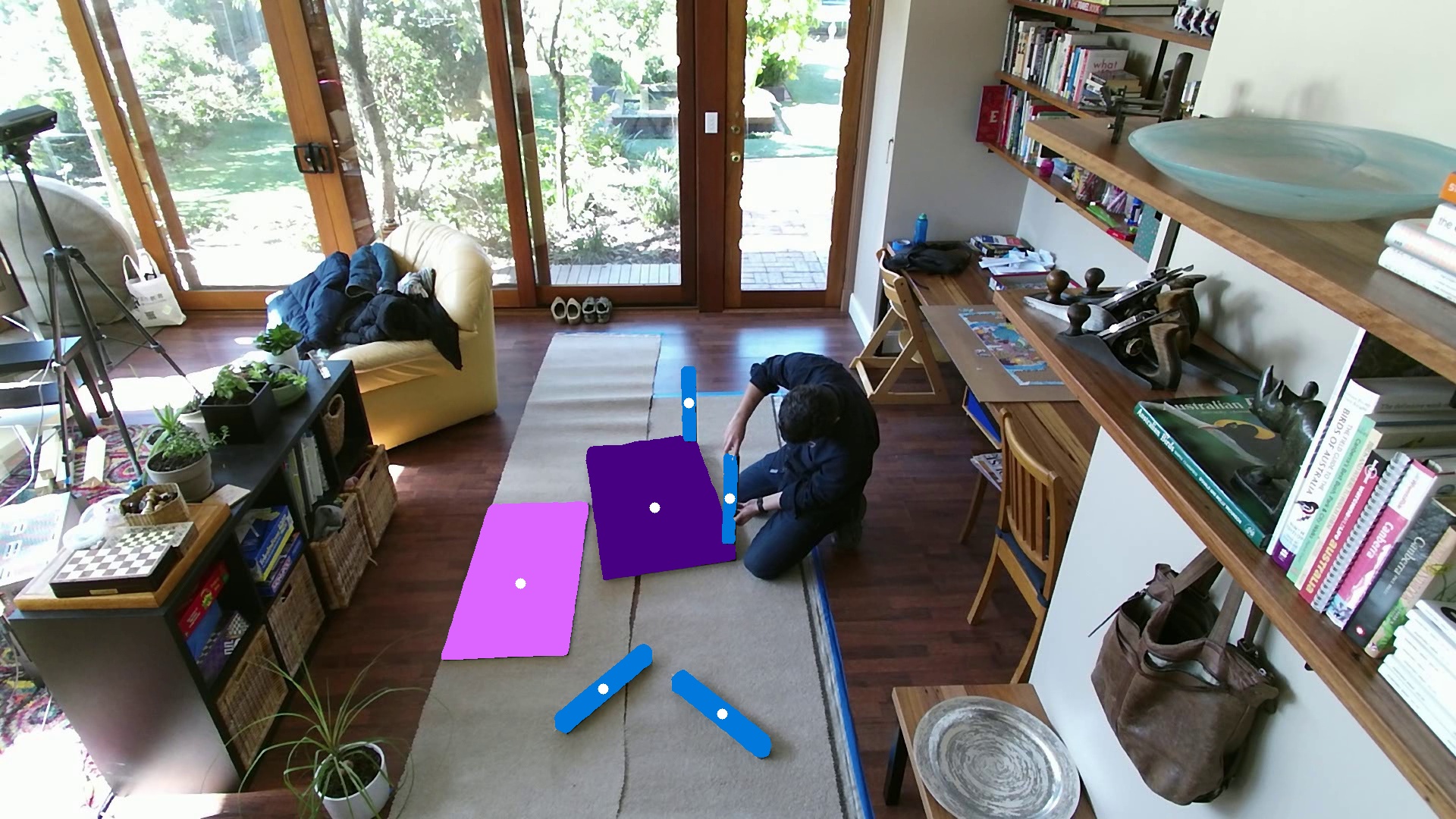}%
	    };%
	    \node[anchor=west] at (6.8, 1.6){\footnotesize{Front View}};%
	    \node[anchor=west] at (5, 0){%
	        \includegraphics[width=4.8cm]{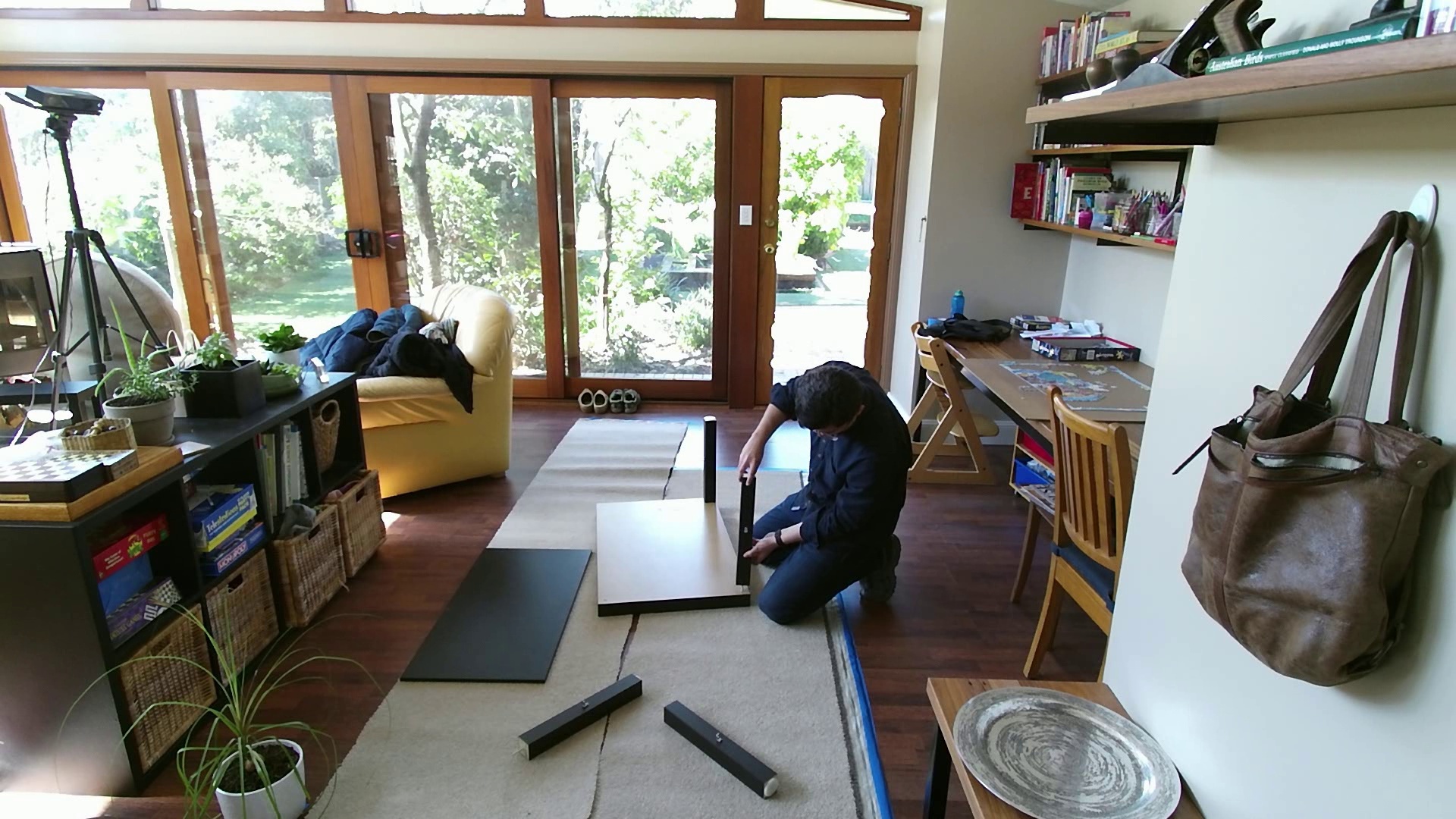}%
	    };%
	    \node[anchor=west] at (11.8, 1.6){\footnotesize{Side View}};%
	    \node[anchor=west] at (10, 0){%
	        \includegraphics[width=4.8cm]{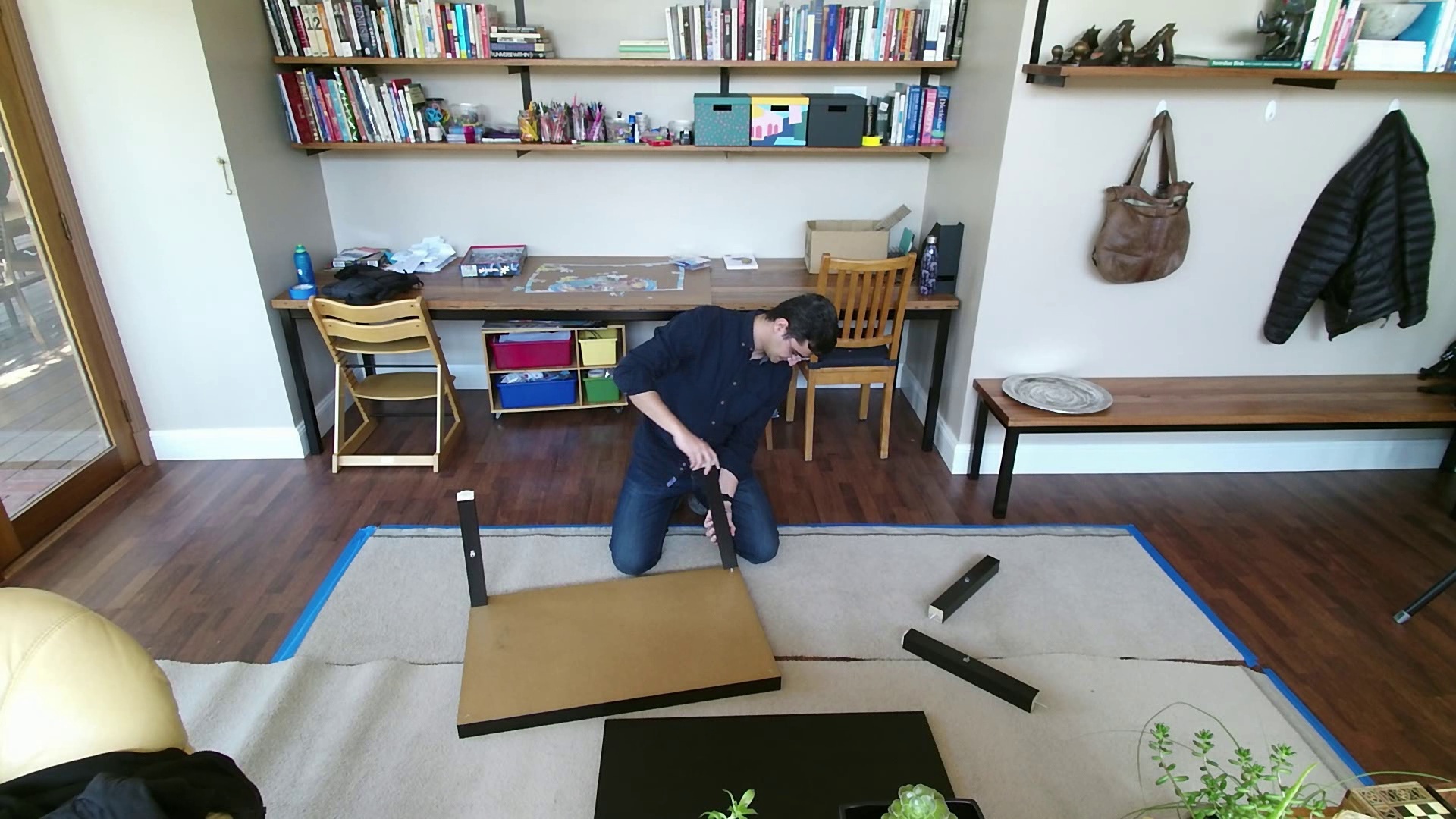}%
	    };%
        \node[anchor=west, rotate=90] at (-0.1, -4.3){\footnotesize{Prediction ($s>0.1$)}};%
	    \node[anchor=west] at (0, -3){%
	        \includegraphics[width=4.8cm]{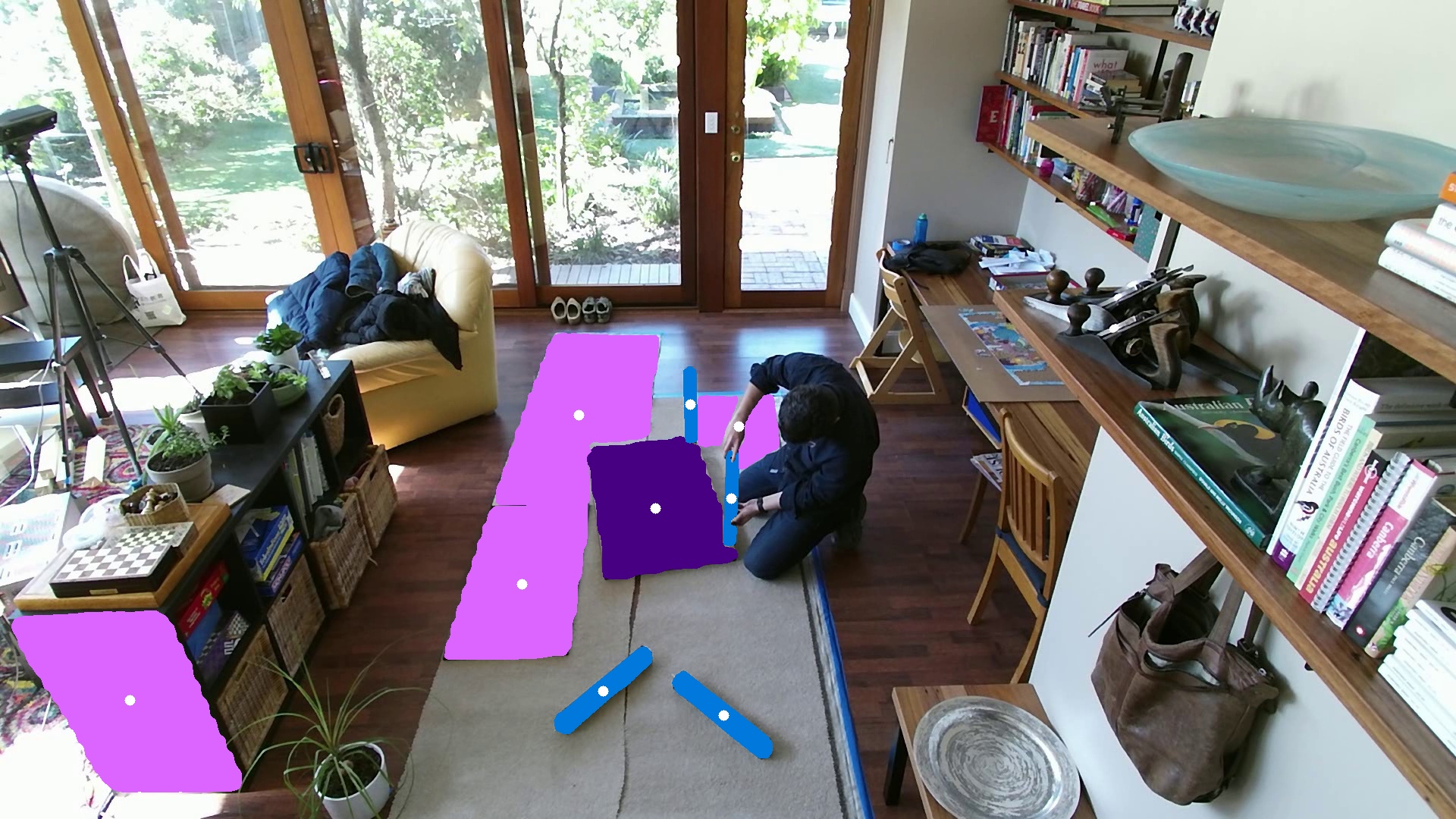}%
	    };%
	    \node[anchor=west] at (5, -3){%
	        \includegraphics[width=4.8cm]{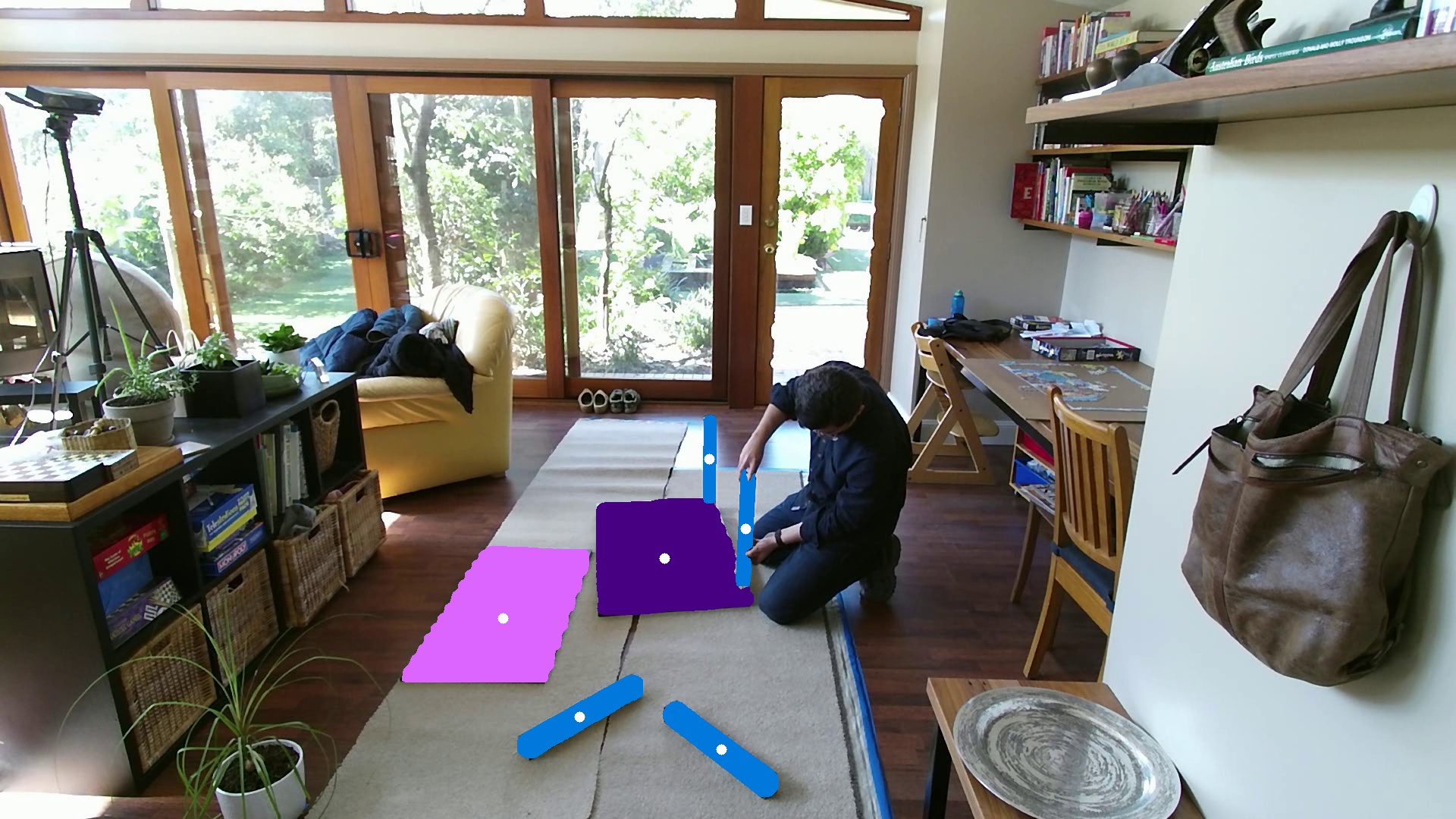}%
	    };%
	    \node[anchor=west] at (10, -3){%
	        \includegraphics[width=4.8cm]{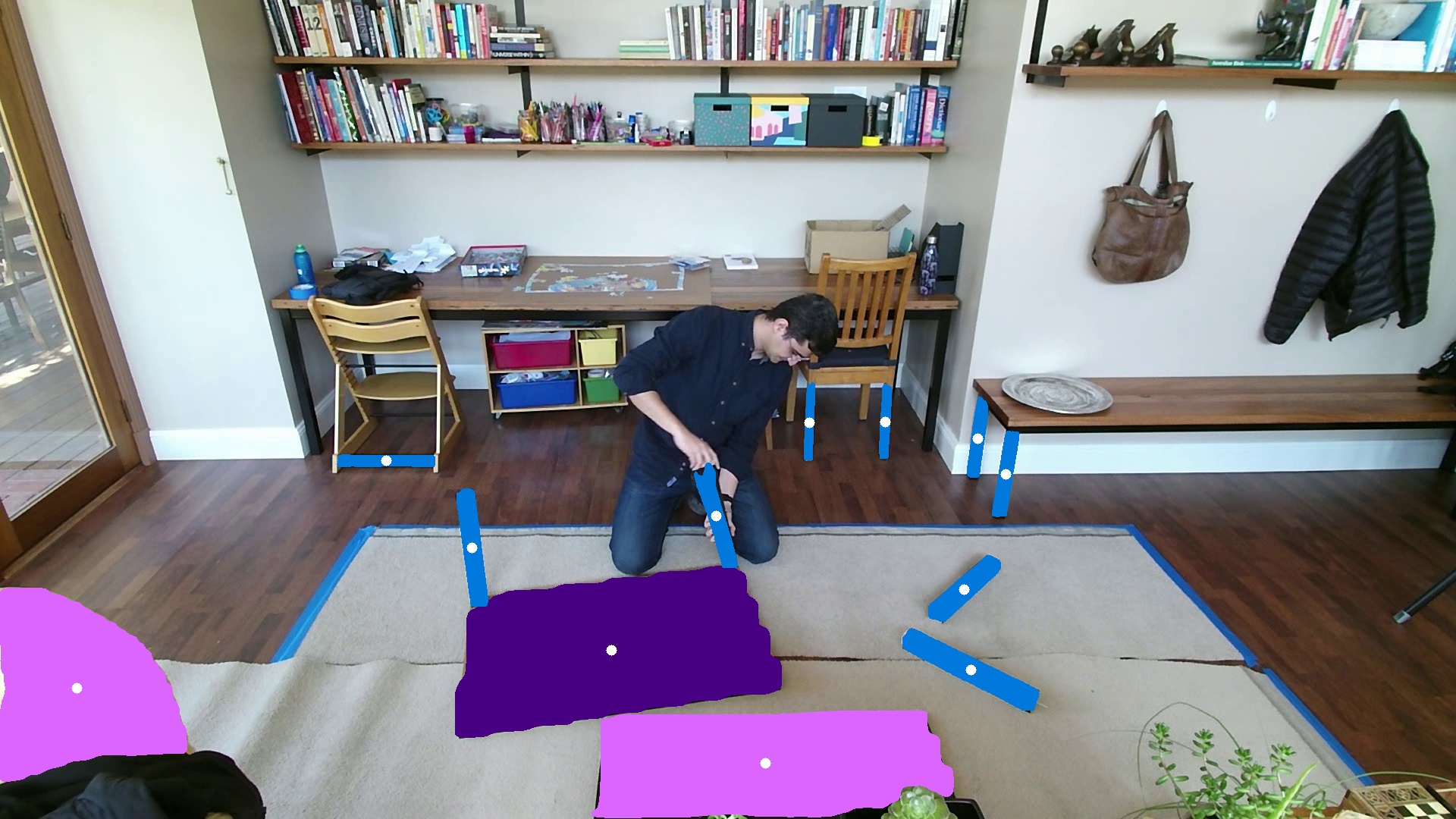}%
	    };%
	    \fill[Table Top] (15.2, -0.0) rectangle (15.6, -0.4);%
        \node[anchor = west] at (15.6, -0.2){\scriptsize{Table Top}};%
        \fill[Leg] (15.2, -0.5) rectangle (15.6, -0.9);%
        \node[anchor = west] at (15.6, -0.7){\scriptsize{Leg}};%
        \fill[Shelf] (15.2, -1.0) rectangle (15.6, -1.4);%
        \node[anchor = west] at (15.6, -1.2){\scriptsize{Shelf}};
        \fill[Side Panel] (15.2, -1.5) rectangle (15.6, -1.9);%
        \node[anchor = west] at (15.6, -1.7){\scriptsize{Side Panel}};%
        \fill[Front Panel] (15.2, -2.0) rectangle (15.6, -2.4);%
        \node[anchor = west] at (15.6, -2.2){\scriptsize{Front Panel}};%
        \fill[Bottom Panel] (15.2, -2.5) rectangle (15.6, -2.9);%
        \node[anchor = west] at (15.6, -2.7){\scriptsize{Bottom Panel}};%
        \fill[Rear Panel] (15.2, -3.0) rectangle (15.6, -3.4);%
        \node[anchor = west] at (15.6, -3.2){\scriptsize{Rear Panel}};%
    \end{tikzpicture}%
    }%
    \caption{
    Instance segmentation of our Mask R-CNN with Swin-Tiny backbone trained on IKEA ASM~\cite{IKEA-wacv2021}.
    The colors of the seven object classes are shown on the right.
    Ground truth labels are only available for the top view.
    We also visualize the center of the object masks with a white dot.
    All predictions with a confidence score $s>0.1$ are shown, as this liberal threshold performed best in our action recognition experiments.
    }
    \label{fig:ikea-asm-objects}
\end{figure*}

IKEA ASM consists of 371 distinct assembly processes, where 48 participants were instructed to assemble one of four different furniture types.
Each assembly process is captured by three different Kinect V2 cameras, resulting in 1113 videos and 35 hours of footage at 24fps.
For human action recognition, there are 17K labeled action instances, distributed over 33 atomic classes.
We use the official splits provided in~\cite{IKEA-wacv2021}, where about $\nicefrac{2}{3}$ is in training and $\nicefrac{1}{3}$ in testing.
The pieces of furniture consist of seven different object categories.
For the object instances 1\% of the data were manually labeled.
The remaining annotations were obtained by overfitting several models on the annotated data, resulting in pseudo ground truth data.
However, the object masks are only available for one view as shown in \autoref{fig:ikea-asm-objects} at the top.
For skeleton data IKEA ASM provides 2D skeleton predictions obtained from OpenPose~\cite{openpose-cvpr2017} and Keypoint R-CNN~\cite{maskrcnn-iccv2017}, of which we use the better performing Keypoint R-CNN in our experiments.

The dataset provides official training and test splits, which we also use in our experiments.

\section{Approach}
\label{section-approach}
For integrating additional object information into skeleton-based action recognition, we build our approach upon two different state-of-the-art methods for skeleton-based action recognition, namely VA-CNN~\cite{VACNN-TPAMI2019} and PoseConv3D~\cite{PoseConv3D-cvpr2022}.

\subsection{VA-CNN Modification}
VA-CNN~\cite{VACNN-TPAMI2019} is a rather fast method while still achieving competitive results.
Different to many other methods, the whole trimmed action sequence is utilized instead of a smaller clip.
The main parts of VA-CNN are a specific input encoding, a view-adaptation module, and a classification network, which we will all describe in the following.

\paragraph{Input Encoding} VA-CNN uses the input encoding from \cite{skeleton-encoding-acpr2015}.
The skeletons for the whole action sequence are transformed into a single RGB image.
One column represents one frame of the action and in each column the skeleton joints are stacked in a fixed order.
The XYZ coordinates of a joint are transformed to RGB by normalizing and scaling:
\DeclarePairedDelimiter\floor{\lfloor}{\rfloor}
\begin{equation}
    u_{t,j} =  255 \cdot \frac{v_{t,j} - c_\text{min}}{c_\text{max} - c_\text{min}} \; ,
\end{equation}
where $v_{t,j}$ denotes the 3D coordinates of the $j^{th}$ joint of the $t^{th}$ frame in a skeleton sequence, $c_\text{min}$ and $c_\text{max}$ are the minimum and maximum of all joint coordinates in the training data, and $u_{t,j}$ refers to the normalized coordinates.
As we work with the 2D skeletons from IKEA ASM~\cite{IKEA-wacv2021}, we only have an image with red and green channels and an empty blue channel.
The resulting output is shown in \autoref{fig:eyecatcher}.
For training and application, each resulting image can then be transformed to the appropriate input size for the classification network using image operations.

\paragraph{View-Adaptation Module} 
The view-adaption (VA) module aims at normalizing the skeletons to one resembling view.
However, we have found that, when using 2D skeletons, this module does not work as intended and has no effect on the final performance.
Therefore, we omit the VA part %
and refer to this modified VA-CNN as 2D-CNN in the remainder of this paper.

\paragraph{Classification Network} 
Here, any classification network can be utilized.
Therefore, we use the popular and reliable \mbox{ResNet-50~\cite{ResNet-cvpr2016}}, which was also used in VA-CNN.

\paragraph{Incorporating Object Information} 
Our general idea for incorporating object information is to consider objects similar to further skeleton joints.
Thus, we combine object data with skeleton data as early as in the input encoding.
We modify the input encoding by appending additional rows below the skeleton joints.
Specifically, we append one additional line for each object class that occurs in the dataset.

Since skeleton joints are represented as points, we also need to convert object masks to this format.
Therefore, we calculate the center of mass for each object mask in order to represent each object as single point as well.
These object coordinates can then be inserted in the appended lines below the skeleton joints. 
As not all object classes appear in all frames, we set the coordinates for the missing classes to zero, as can be seen by the black parts in \autoref{fig:eyecatcher}.

Unfortunately, this encoding for 2D-CNN has a drawback.
Since a human typically has only one of each type of skeleton joints, the encoding was not designed to include multiple instances of the same joint class.
Thus, we are presented with a challenge when dealing with varying numbers of objects in each sequence or sometimes even each frame.
For example, during assembling a table, there are multiple table legs, but only one of these legs can be encoded.
Even if each object class occurs only once, this issue can still occur when working with a real object detector, which also produces false positives that cannot be filtered completely by tuning the confidence threshold.

Therefore, we decided to only use one object coordinate of each object class.
This leaves us with the question which object to use if there are multiple instances of the same class.
To overcome this challenge, we use a simple heuristic:
The objects a human interacts with are more relevant to action recognition than others.
For interacting with objects, humans typically use their hands.
Therefore, we calculate the euclidian distance of all object coordinates to the predicted hand joints, and choose the object with the lowest distance to both hand joints for each class.
Alternatively, we also experimented with the lowest distance to only one hand joint, which led to slightly worse results.
As we can see in \autoref{fig:ikea-asm-objects}, our heuristic chooses the table leg that is most important for the current assembly step.

In our experiments, we show that our enhancement for incorporating object information results in an exceptionally high boost in action recognition performance.

\subsection{PoseConv3D Modification}
In contrast to VA-CNN, PoseConv3D~\cite{PoseConv3D-cvpr2022} is a more complex method, using 3D convolutions, while processing 2D skeletons.
Moreover, PoseConv3D is able to process multiple instances of the same skeleton joint class, since it can handle multiple detected human skeletons, thereby processing the joints of the group of persons.

For PoseConv3D the input encoding is also the most interesting part:
The skeleton joints of a single frame are embedded in heatmaps -- one heatmap for each joint class -- which are stacked along the channel axis, resulting in a heatmap tensor of size $J \times H \times W$, where $J$ is the number of distinct joint classes and $H$ and $W$ refer to the height and width of the cropped frame.
The heatmaps are obtained as following:
The $j^{th}$ joint coordinate $(c_x^{(j)}, c_y^{(j)}, s^{(j)})$ with the location $(c_x^{(j)}, c_y^{(j)})$ and the confidence score $s^{(j)}$ is embedded in a heatmap $K$ with a Gaussian kernel centered at the joint:
\begin{equation}
    K_{j,x,y} = e^{-\frac{\left(x-c_x^{(j)}\right)^2 + \left(y-c_y^{(j)}\right)^2}{2 \cdot \sigma^2}} \cdot s^{(j)},
\end{equation}
where $\sigma$ denotes the variance of the Gaussian and is set according to the original PoseConv3D implementation~\cite{PoseConv3D-cvpr2022}.
For each joint, the multiple heatmaps over time are then stacked in an additional time axis $T$, resulting in a 4-dimensional input $x \in \mathbb{R}^{J \times T \times H \times W}$, which is illustrated in \autoref{fig:eyecatcher}.
This input is then fed into the 3D-CNN SlowOnly~\cite{slowonly-iccv2019}, which is directly inflated from ResNet-50.

Similar to our 2D-CNN method, we can append object information to the skeletons, here by using additional heatmaps.
For this, we again consider the object centers as further skeleton joints and append one additional heatmap for every object class.
In contrast to 2D-CNN, PoseConv3D offers the major advantage that it can handle multiple instances of the same joint class.
This means that we are not restricted to one single object per class, but can simply draw a Gaussian for each object of the same class in one single heatmap.
Likewise, for detected object masks, we have the advantage of also modeling the estimated confidence directly with the Gaussian.

\vspace{3mm}
In order to prove the usefulness of object information for human action recognition, we experiment with both ground truth as well as detected object masks from an instance segmentation model.

\section{Setup}
\label{section-experiments-without-objects}

For our implementation, we use the publicly available code from VA-CNN\footnote{\url{https://github.com/microsoft/View-Adaptive-Neural-Networks-for-Skeleton-based-Human-Action-Recognition}} for our 2D-CNN method and MMAction2~\cite{2020mmaction2} for PoseConv3D.
Overall, we followed the original pipelines and hyperparameters and only adapted the following:
For training, we use the Adam optimizer~\cite{kingma2014adam} and a OneCycle learning rate scheduler with a 10\% warmup and different maximum learning rates.
In general, we always perform a linear learning rate search close to the original learning rates and repeat each configuration two to three times to obtain consistent results.

We evaluate the performance of our models using mean class accuracy (mAcc) as well as top-1 accuracy (top1) as these are commonly used for pre-trimmed sequences in many action recognition papers~\cite{assembly101-cvpr2022, Carreira2017Kinetics, ntu-tpami2019, meccano-wacv2021}.
For the top1 metric, all action clips contribute equally.
In contrast, mAcc averages accuracy across all action classes, which compensates for class imbalances.

Our experiments are sectioned into two parts:
First, in \autoref{section-experiments-ground-truth}, we analyze the impact of integrating additional object information on human action recognition using ground truth data.
Afterwards, in \autoref{section-experiments-prediction}, we train our own instance segmentation model and use the resulting object mask predictions in combination with skeletons for action recognition.

\section{Experiments with Ground Truth Object Masks}
\label{section-experiments-ground-truth}
In the following, we analyze the extent to which combining object centers with skeleton joints can improve skeleton-based human action recognition.
In order to demonstrate the general viability of our approach and to be independent from the performance of an object detector, we first use the ground truth data of IKEA ASM for objects (\autoref{subsection-experiments-ground-truth}).
We also present an ablation study on a modified IKEA ASM dataset to investigate the benefits of the combination, even if the object classes have a small influence on the classes to be predicted (\autoref{subsection-experiments-only-verbs}).
The subsequent section \autoref{section-experiments-prediction} then presents our experiments with object masks, determined by an instance segmentation model.

\subsection{Preliminary Considerations for Using Ground Truth \mbox{Objects}}
When training with ground truth data for object information, we are faced with the issue that the IKEA ASM dataset has three views, but objects are only annotated in the top view as shown in \autoref{fig:ikea-asm-objects}.
To solve this issue, we considered the following solutions:
\paragraph{Only Use Top View Data}
One option could be to only use the camera view on which the objects have been labeled and discard the other two views.
However, this leads to a smaller and simpler data subset of only one third.
Furthermore, it limits comparability to predicted object masks in \autoref{section-experiments-prediction}.

\paragraph{Use All Views, but Add Objects Only to Top View}
Another option could be to use skeleton data from all views and combine them with object data only in case of the top view.
This allows us to compute results over the whole test set, leading to better comparability.
However, as $\sfrac{2}{3}$ of the dataset would contain no object information at all, the impact of additional object information might be limited and it might be hard to asses the actual potential. %

\paragraph{Add Object Centers to Every Skeletal Perspective} 
In order to increase the impact of additional object information, we could also add the object centers from the top view to all views.
This seems counterintuitive at first, as the object center coordinates do not match the skeleton joint coordinates for front and side view.
As we are only dealing with 2D data and have neither depth information nor extrinsics for all three views, we are also unable to transform the data to a corresponding perspective.
Yet, in preliminary experiments with 2D-CNN, we found that option c) leads to the best results, i.e., $47.8$ mAcc, compared to only $37.5$ mAcc for option b).
This already shows how important object information seems to be, if the mAcc is up to $10$ percentage points worse, if $\sfrac{2}{3}$ of the training and test data consisted of skeletal data only.
Therefore, we used option c) for all experiments in this section, enabling us to evaluate the potential of combining object information with skeleton joints for human action recognition.%
\footnote{As a side note: The combination of skeleton data and object data according to option a), i.e., using only the top view for training and testing, the 2D-CNN method achieved a mAcc of $46.6\%$ on the limited test dataset.}

\subsection{Assessing the Potential of Skeleton-Object-Combination}
\label{subsection-experiments-ground-truth}
\paragraph[Skeleton Only Baseline]{\skeleton ~-- Skeleton Only Baseline} To establish a general baseline, we first trained both \mbox{2D-CNN} and PoseConv3D using \emph{only skeletons}.
As we can see in the first row of \autoref{tab:baselines}, the much more complex PoseConv3D is also significantly better on only skeleton sequences, so the results from the state of the art also apply to the IKEA ASM dataset.

\begin{table}[htb]
\centering
\caption{Results using Ground Truth Object Masks on IKEA ASM~\cite{IKEA-wacv2021}}
\label{tab:baselines}
\begin{tabular}{cccc@{\hspace{2mm}}c@{\hspace{6mm}}c@{\hspace{2mm}}c}
\toprule
          &         &                      & \multicolumn{2}{c}{\textbf{2D-CNN\phantom{iiii}}} & \multicolumn{2}{c}{\textbf{PC3D\phantom{iii}}} \\
\skeleton & \object & used objects         & mAcc             & top1                           & mAcc            & top1                         \\ \midrule
\yes      &         &                      & 37.7             & 70.3                           & 39.5            & 75.0                         \\\rule{0pt}{4mm}%
          & \yes    & \emph{most relevant} & 44.0             & 67.0                           & 45.8            & 75.0                         \\
          & \yes    & \emph{all}           & ---              & ---                            & 51.7            & 78.5                         \\\rule{0pt}{4mm}%
\yes      & \yes    & \emph{most relevant} & \textbf{47.8}    & \textbf{79.6}                  & 56.5            & 84.3                         \\
\yes      & \yes    & \emph{all}           & ---              & ---                            & \textbf{58.5}   & \textbf{85.4}                \\ \bottomrule
\end{tabular}%

\end{table}

\paragraph[Skeleton-Object-Combination]{\skeletonAndObject ~-- Skeleton-Object-Combination}
For analyzing the combination of objects and skeletons, it must be taken into account that there are multiple ways to combine object information with skeleton joints, depending on the used methods 2D-CNN and PoseConv3D (as described in \autoref{section-approach}).
Therefore, we carried out various experiments in order to be able to compare the methods as fair as possible.

For our 2D-CNN method, we are restricted to only one object per class, which we will refer to as the \emph{most relevant objects} in the following.
In contrast, for PoseConv3D this restriction does not apply.
Thus, first we trained PoseConv3D with \emph{all objects} in every frame and second also only with the \emph{most relevant objects} for better comparability to 2D-CNN.
\autoref{tab:baselines} presents the results of these experiments.

Compared to using only skeletons, the combination with object information improves the performance to a large extent for both methods.
For 2D-CNN, we can gain about 10 percentage points mAcc.
For PoseConv3D the improvement is even greater with 19 percentage points mAcc.
As expected, PoseConv3D performs better when it has access to all object data.
These results show how important object information is for skeleton-based action recognition when the acting person interacts with objects.
And thus, these results also show that purely skeleton-based inputs are severely disadvantaged in such action recognition tasks and that fundamentally important information is typically lost when using only skeleton sequences.

\paragraph[Only Ground Truth Object Information]{\object ~-- Only Ground Truth Object Information}
Due to this large performance gain, we decided to investigate how well the action classifier performs when we train our models using only ground truth object information, and consequently omit skeleton data.
The results are shown in the middle rows of \autoref{tab:baselines}.
Surprisingly, the models trained only on ground truth object information show a very large performance gain compared to the models trained only on skeleton information.
This result shows how much a perfect object detector can contribute to robust action recognition.
Nevertheless, it must be taken into account that this comparison is not completely fair here, since the object models were trained and tested exclusively on the top view perspective.
A more fair comparison in which we only use real predicted object information on all perspectives is presented in \autoref{section-experiments-prediction}.

Furthermore, we have to keep in mind, that the actions of the IKEA ASM dataset such as \textit{pick up leg} or \textit{pick up shelf} can be much better distinguished by moving objects than by a moving skeleton.
Therefore, we also carried out an ablation study on a dataset with combined action classes, which we will describe in the following section.

\subsection{Verb-focused Action Recognition Study}
\label{subsection-experiments-only-verbs}
The IKEA ASM dataset contains classes for action recognition, which can only be distinguished by the object used, such as: \textit{attach back panel}, \textit{attach side panel}, or \textit{attach shelf}.
For the subject area of human-object interaction recognition, this question is highly relevant, but for some applications, such as general human action recognition, it is sufficient to know only the action in general, i.e., the verb describing the action.
In order to investigate the usefulness of combining object information with skeleton joints for such application domains, we modify the IKEA ASM dataset such that the action classes to be predicted are no longer so heavily dependent on the objects used.
Thus, we summarize the 33 action classes based on the verbs, resulting in an \emph{only verbs} version of the IKEA ASM dataset.
This leads to 12 remaining action classes (\textit{align}, \textit{attach}, \textit{flip}, \textit{insert}, \textit{lay down}, \textit{pick up}, \textit{position}, \textit{push}, \textit{rotate}, \textit{slide}, \textit{spin}, \textit{tighten}).

Subsequently, we repeated experiments conducted above on the \emph{only verbs} dataset and present results in \autoref{tab:verbs}.
Note that these results cannot directly be compared to the results in other tables, as the \emph{only verbs} dataset consists of combined action classes, leading to fewer confusions between different classes.

\begin{table}[htb]
\centering
\caption{Results on our Only Verbs Version of IKEA ASM~\cite{IKEA-wacv2021}}
\label{tab:verbs}
\begin{tabular}{cccc@{\hspace{2mm}}c@{\hspace{6mm}}c@{\hspace{2mm}}c}
\toprule
          &         &                      & \multicolumn{2}{c}{\textbf{2D-CNN\phantom{iiii}}} & \multicolumn{2}{c}{\textbf{PC3D\phantom{iii}}} \\
\skeleton & \object & used objects         & mAcc             & top1                           & mAcc            & top1                         \\ \midrule
\yes      &         &                      & 63.7             & 82.6                           & 64.7            & 85.8                         \\\rule{0pt}{4mm}%
          & \yes    & \emph{most relevant} & 63.8             & 78.3                           & 74.2            & 83.6                         \\
          & \yes    & \emph{all}           & ---              & ---                            & 77.4            & 88.9                         \\\rule{0pt}{4mm}%
\yes      & \yes    & \emph{most relevant} & \textbf{69.4}    & \textbf{86.7}                  & \textbf{81.4}   & 91.5                         \\
\yes      & \yes    & \emph{all}           & ---              & ---                            & 81.1            & \textbf{92.0}                \\ \bottomrule
\end{tabular}%
\end{table}

First, generally speaking, the same trends can be seen as on the original IKEA ASM dataset.
The more computationally intensive PoseConv3D is generally always better than the 2D-CNN.
Likewise, the combination of skeleton joints and object information again shows to be enormously beneficial.
The mAcc for 2D-CNN is improved by 6 percentage points and for PoseConv3D by 17 percentage points compared to training only on skeleton joints.
While using only objects results in similar mAcc performance to using only skeletons for \mbox{2D-CNN}, we are still observing a large performance gain of 13 percentage points for PoseConv3D.

Therefore, we could effectively demonstrate that the performance gain of incorporating object information cannot only be attributed to the way the action classes are defined in IKEA ASM dataset.
On the contrary, even for the verbs-only action classes, object information still increases performance to a large extent.

\section{Experiments with Predicted Object Masks}
\label{section-experiments-prediction}

In the previous section, we presented the potential of combining object information with skeleton joints by utilizing the ground truth object masks. %
For the following experiments, we first train an instance segmentation model (\autoref{subsection-instance-segmentation-experiments}).
Afterwards, we combine its predicted object masks with skeleton joints for human action recognition (\autoref{subsection-action-recognition-pred-obj}).
Thus, we show the impact of using non-perfect predicted object masks, which typically also contain false positive as well as false negative predictions.

As mentioned before, so far we could only use object coordinates from top view, since there are no labels for the other two views.
In this section, we will counteract this issue by applying our trained instance segmentation model to predict object masks for all views.
This allows us to combine object and skeleton information from corresponding perspectives.
Moreover, we can investigate how non-matching object and skeletal perspectives might have negatively affected the learning process, leading to potentially worse performance.

\subsection{Instance Segmentation}
\label{subsection-instance-segmentation-experiments}
\begin{table}[b]
    \centering
    \caption{Performance of Mask R-CNN instance segmentation models with different backbones on IKEA ASM~\cite{IKEA-wacv2021}.}
    \label{tab:instance-segmentation}
    \begin{tabular}{lcr@{\hspace{2mm}}r@{\hspace{6mm}}r@{\hspace{2mm}}r}
      \toprule
                                                &                      & \multicolumn{2}{c}{mask\phantom{M}}  & \multicolumn{2}{c}{bbox\phantom{m}} \\
      backbone                                  & weights              & AP                   & AP50          & AP            & AP50                \\
      \midrule
      ResNet-50 \cite{ResNet-cvpr2016}          & \cite{IKEA-wacv2021} & 58.1                 & 77.2          & 59.5          & 77.7                \\
      ResNet-101 \cite{ResNet-cvpr2016}         & \cite{IKEA-wacv2021} & 62.1                 & 82.0          & 64.6          & 81.8                \\
      ResNeXt-101 \cite{resnext-cvpr2017}       & \cite{IKEA-wacv2021} & \textbf{65.9}        & 85.3          & 69.5          & 86.4                \\
      Swin-Tiny \cite{swin-transfomer-iccv2021} & ours                 & 63.9                 & \textbf{89.4} & \textbf{72.5} & \textbf{92.1}       \\
      \bottomrule
    \end{tabular}
\end{table}
For training an instance segmentation model on the IKEA ASM dataset, the 1\% manually labeled frames from top view are used.
The authors of the IKEA ASM dataset~\cite{IKEA-wacv2021} already provide pretrained weights for Mask R-CNN~\cite{maskrcnn-iccv2017} with ResNet-50, ResNet-101 as well as ResNeXt-101 backbone.
Moreover, we also train our own Mask R-CNN with a Swin Transformer Tiny~\cite{swin-transfomer-iccv2021} backbone, as the Swin Transformer is a state-of-the-art model outperforming ResNet in many applications.
For training on the IKEA ASM dataset, we used a model pretrained on MS COCO~\cite{coco-eccv2014} and finetuned it for 12 epochs using the AdamW optimizer~\cite{adamw-iclr2019} and a OneCycle learning rate scheduler with a maximum learning rate of $5\cdot10^{-5}$.
While having similar complexity with ResNet-50, our Swin-Tiny Mask R-CNN reaches an even higher performance than the ResNeXt-101 model in most metrics, as shown in \autoref{tab:instance-segmentation}.
Qualitative results of our trained model on all three views are shown in the second row of \autoref{fig:ikea-asm-objects}.
Overall, we observed very few false negative predictions, but -- depending on the confidence threshold -- several false positive predictions.

After training our model, we predicted object masks for all frames of all views.
We saved all predictions with a confidence score $s>0.1$, which allows us to later experiment with different thresholds for human action recognition.

\subsection{Human Action Recognition}
\label{subsection-action-recognition-pred-obj}

In the following, we first describe the main motivation of the series of experiments in this section and the resulting different training settings we need to consider.
We also describe how we choose which of the dedicated object masks to use for training and testing, and the different resulting training configurations.
Finally, we conclude with the experimental results and compare them to our baseline, as well as the results on ground truth object masks.

\paragraph{Preliminary Considerations}
The purpose of the following experiments on human action recognition is twofold and thus we have to consider two different training settings.
\begin{enumerate}
    \item \emph{Top View:}
    On the one hand, we want to evaluate the potential performance drop of using predicted instead of ground truth object masks.
    In order to compare these results with our experiments in \autoref{section-experiments-ground-truth}, we also use predicted object masks only for the top view, but combine them with skeletons from the other two views as well.
    \item \emph{All Views:}
    On the other hand, we want to investigate the benefit of combining real predicted object masks and skeleton joints for action recognition closer to a real-world application.
    Thus, the perspective of the predicted objects mask should always match the perspective of the skeleton joints.
    This should also make it easier for the models to combine the data appropriately and facilitate the learning process.
\end{enumerate}

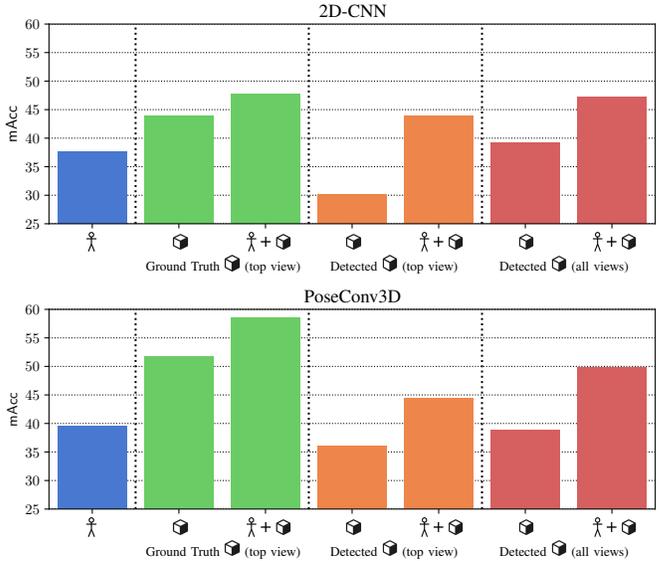
\begin{figure}[bt]
   \centering
    \begin{tikzpicture}
        \node[anchor=west] at (0, 0){%
	        \includegraphics[width=\columnwidth, trim={0mm 0mm 0mm 0mm}]{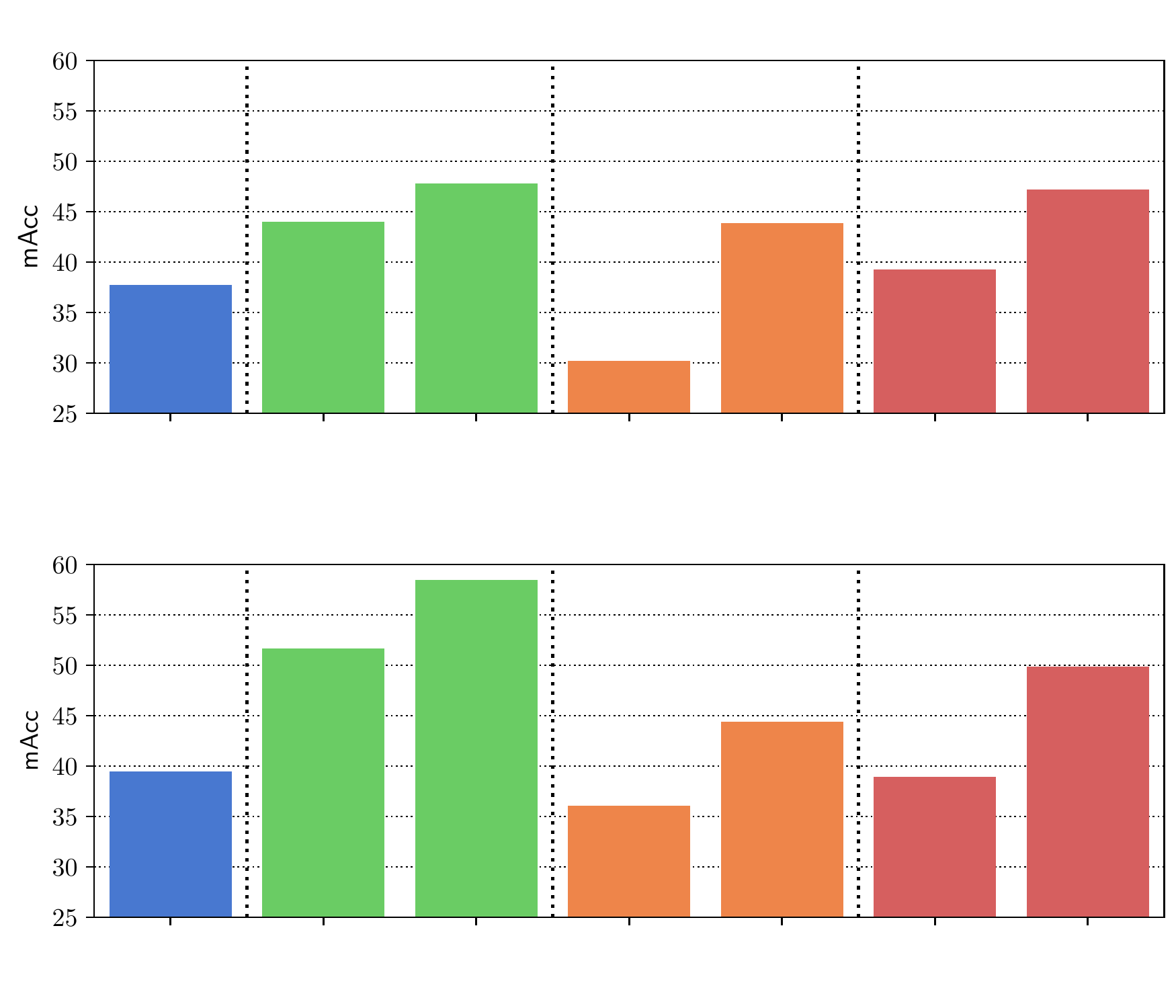}%
	    };%
        \node[anchor=west] at(4.3, 3.52){\scriptsize{2D-CNN}};
        \node[anchor=west] at(1.2, 0.44){\skeleton};
        \node[anchor=west] at(2.35, 0.44){\object};
        \node[anchor=west] at(3.33, 0.44){\skeletonAndObject};
        \node[anchor=west] at(4.65, 0.44){\object};
        \node[anchor=west] at(5.63, 0.44){\skeletonAndObject};
        \node[anchor=west] at(6.95, 0.44){\object};
        \node[anchor=west] at(7.93, 0.44){\skeletonAndObject};
        \node[anchor=west] at(2.0, 0.14){\tiny{Ground Truth \object ~(top view)}};
        \node[anchor=west] at(4.45, 0.14){\tiny{Detected \object ~(top view)}};
        \node[anchor=west] at(6.7, 0.14){\tiny{Detected \object ~(all views)}};
        \node[anchor=west] at(4.1, -0.28){\scriptsize{PoseConv3D}};
        \node[anchor=west] at(1.2, -3.35){\skeleton};
        \node[anchor=west] at(2.35, -3.35){\object};
        \node[anchor=west] at(3.33, -3.35){\skeletonAndObject};
        \node[anchor=west] at(4.65, -3.35){\object};
        \node[anchor=west] at(5.63, -3.35){\skeletonAndObject};
        \node[anchor=west] at(6.95, -3.35){\object};
        \node[anchor=west] at(7.93, -3.35){\skeletonAndObject};
        \node[anchor=west] at(2.0, -3.65){\tiny{Ground Truth \object ~(top view)}};
        \node[anchor=west] at(4.45, -3.65){\tiny{Detected \object ~(top view)}};
        \node[anchor=west] at(6.7, -3.65){\tiny{Detected \object ~(all views)}};
    \end{tikzpicture}
    \vspace*{-6mm}
    \caption{
    Comparison of action recognition performance on IKEA ASM~\cite{IKEA-wacv2021} with \skeleton: only skeletons, \object: only objects and \skeletonAndObject: the combination of both.
    }
    \label{fig:performance-comparison}
\end{figure}

\paragraph{Selecting the Detected Object}
For 2D-CNN, we are restricted to one object per class, which we determine by calculating the distance to the hand joints.
However, different confidence thresholds for the predicted objects lead to different chosen objects.
Therefore, we performed an equidistant line search to determine the best confidence threshold for all experiments.
Overall, a threshold of $\tau_s=0.1$ led to the best results, and we present all results using this threshold.
We assume that the distance to the hand joints is a good heuristic for obtaining the most relevant object, even when the predictions contain several false positives as shown in \autoref{fig:ikea-asm-objects}.
We therefore hypothesize that higher thresholds with more false negatives (i.e., more missing objects) have a greater adverse effect than smaller thresholds with more false positives.
In the following experiments, models trained with this kind of object pre-selection are again marked as \emph{most relevant} objects used.

For PoseConv3D, this restriction does not apply.
We are able to use all objects, and scale the maximum values in the heatmaps by the predicted confidence scores.
Implicitly, this should give the model the intuition that objects with a higher confidence value might be more relevant.
Experimental results from this training setting are consequently marked as \emph{all} objects used.
However, for comparison reasons we also experiment on PoseConv3D with \emph{most relevant} objects used.

We use the predicted object masks from Mask R-CNN with Swin-Tiny backbone in all of our experiments in this section.
As the results on instance segmentation in \autoref{tab:instance-segmentation} have already indicated, using predictions from other models led mostly to worse action recognition results in preliminary experiments and for the more computationally expensive model with ResNext-101 backbone to no clear advantage.

\paragraph{Results with Detected Object Masks}
We present the results for the settings described above for our experiments of combining detected object masks and skeleton joints in \autoref{tab:results_pred_obj_masks}.

\begin{table}[!b]
\centering
\caption{Results using Detected Object Masks on IKEA ASM~\cite{IKEA-wacv2021}}
\label{tab:results_pred_obj_masks}
\begin{tabular}{lcccc@{\hspace{2mm}}c@{\hspace{6mm}}c@{\hspace{2mm}}c}
\toprule
                                                                         & \multicolumn{1}{l}{} & \multicolumn{1}{l}{} & \multicolumn{1}{l}{} & \multicolumn{2}{c}{\textbf{2D-CNN\phantom{iiii}}} & \multicolumn{2}{c}{\textbf{PC3D\phantom{iii}}} \\
                                                                         & \skeleton            & \object              & used objects         & mAcc                    & top1                    & mAcc                   & top1                  \\ \midrule
\multirow{2}{*}{\rotatebox[origin=c]{90}{\parbox{0.5cm}{~top\\ view}}}   & \yes                 & \yes                 & \emph{most relevant} & 43.9                    & 75.6                    & 44.4                   & 75.1                  \\
                                                                         & \yes                 & \yes                 & \emph{all}           & ---                     & ---                     & 41.6                   & 72.1                  \\ \midrule\rule{0pt}{3mm}%
\multirow{2}{*}{\rotatebox[origin=c]{90}{\parbox{0.5cm}{~~all\\ views}}} & \yes                 & \yes                 & \emph{most relevant} & \textbf{47.2}           & \textbf{77.7}           & 49.0                   & \textbf{80.2}         \\
                                                                         & \yes                 & \yes                 & \emph{all}           & ---                     & ---                     & \textbf{49.9}          & 79.7                  \\ \bottomrule
\end{tabular}%
\end{table}

Likewise, in \autoref{fig:performance-comparison} we present these results in comparison to our skeleton only baseline and the experiments with the combination of ground truth object masks.
For a better overview, this figure shows only the best results of \mbox{PoseConv3D} regarding the setting which objects were used.

Examining these results, the obvious first observation is again that PoseConv3D performs better than 2D-CNN on the combination of skeleton joints and objects.
However, the difference is nowhere near as large as in the previous comparisons.
When comparing the skeleton-only baseline and the combination of skeleton joints with detected object masks, a large performance gain can be seen in \autoref{fig:performance-comparison}.
On \emph{all views}, for example, both the 2D-CNN and the PoseConv3D are about 10 percentage points better on the mAcc.
This shows that our modified methods can handle imperfectly predicted object masks well, and thus we highly encourage to also combine skeletons with predicted object masks for human action recognition in real-world applications.

When comparing the results with ground truth object masks and the results with predicted object masks, we need to look at the experiments on \emph{top view} respectively.
Here we can see that the models produce worse results, with the PoseConv3D showing a much greater performance drop than the 2D-CNN.
This shows that our instance segmentation model, which achieves a very high segmentation AP50 of $89.4$, still has significant weaknesses in predicting object masks, and that these false detections make action recognition significantly more difficult compared to the usage of ground truth object masks.

However, our instance segmentation model offers a major benefit in that it can predict object masks for all perspectives.
When comparing the results of combining predicted object masks and skeleton joints for \emph{top view} and \emph{all views}, we clearly see a considerable improvement.
The 2D-CNN effectively utilized predicted object masks on \emph{all views} and skeleton joints to achieve results that are on par with those obtained with ground truth object masks.
This indicates that the influence of noisy predictions can be better compensated by the models when the perspective of the predictions fits the perspective of the skeleton joints.

The impact of noisy predictions is also shown by the PC3D results when comparing the difference between \emph{most relevant} and \emph{all} objects over \emph{top view} and \emph{all views}.
When training with less data \emph{(top view)}, it has shown to be necessary to filter the \emph{most relevant} objects to handle the noisy predictions.
When training on more data \emph{(all views)}, the model learns to compensate for the noisy predictions by itself.

Finally, we examine the results utilizing only object masks without skeletons.
As expected, the models consistently achieve worse results than with ground truth object masks.
However, we can also see that the results for predicted object masks are significantly better on \emph{all views} compared to only using \emph{top view}.
Again this shows that the models trained on \emph{all views} benefit from more diverse training data from different perspectives to compensate for the noisy predictions.

\paragraph{Training on Ground Truth and Testing on Detected Object Masks}
For human action recognition, one could also consider training with ground truth objects and then apply the trained action recognition model and test with predicted objects.
However, we want to emphasize that this setting leads to considerably worse results.
Our \mbox{2D-CNN} trained with ground truth object masks and tested with predicted object masks (both from \emph{top view}) achieves a mAcc of only 40.2, which is 4 percentage points worse than if we had trained directly with predicted object masks from \emph{top view}.
These results show that the performance drops considerably, if the object detector used in a real-world application does not match the object detector used for training.
This is consistent with literature, where this phenomenon is described as sudden concept drift between training and deployment~\cite{lu2019learning,bayram2022concept}.

\section{Conclusion}
In this paper, we extended two state-of-the-art methods for skeleton-based action recognition to be able to process object information in addition to skeleton joints.
Using these modified methods, we then performed investigations on the challenging assembly dataset IKEA ASM~\cite{IKEA-wacv2021}.
We evaluated the benefit of object information for action classification of human-object interaction classes, as well as for verb-based action classification only.
We also showed the importance of matching the perspective of objects to skeletons and examined the influence of noisy predictions of an object detector on the performance of action classification.

Overall, incorporating additional object information into skeleton-based action recognition improves the performance to a large extent.
Our proposed approach works extremely well for the two considered state-of-the-art skeleton-based action recognition methods 2D-CNN and PoseConv3D.
Moreover, we assume that our approach of treating objects similar to further skeleton joints can also be transferred to other end-to-end deep-learning-based methods for skeleton-based action recognition.

However, the combination of object and skeleton information is a relatively new area of study with ample opportunities for future exploration.
For example, one could also think of using complete object masks, instead of the center of mass, e.g., the input encoding from PoseConv3D would be very suitable for this.
Moreover, instead of combining the inputs, one could also use a dual-branch network and fuse skeleton and object information in the network.
Furthermore, this approach could also be applied to other applications besides assembly, where object information also plays a great role, such as cooking applications (e.g. on the dataset EPIC-KITCHENS-100~\cite{Damen2022EpicK100}).
Perhaps it could even improve action recognition for more general applications, for which the Kinetics400 dataset~\cite{Carreira2017Kinetics} could be utilized in combination with a general object detector trained on MS COCO~\cite{coco-eccv2014}.

With our findings, we hope to have paved the way for more researchers to experiment with combining object information and skeleton data for human action recognition.
We expect that more research in this area enables cobots to recognize human actions more robustly and accurately.
This would improve human-robot collaboration, which is one of the major goals of industry 4.0.

\bibliographystyle{IEEEtran}
\bibliography{literature.bib}

\end{document}